\documentclass[10pt,a4paper]{article}

\usepackage{lrec2000}

\usepackage{linguex}
\usepackage{epsfig}
\usepackage{hyperref}

\title{Annotating Predicate-Argument Structure for a Parallel Treebank}

\name{Lea Cyrus, Hendrik Feddes, Frank Schumacher}

\address{%
  Arbeitsbereich Linguistik, University of M\"{u}nster\\
  H\"{u}fferstra{\ss}e 27, 48149 M\"{u}nster, Germany\\
  \{lea,feddes,frank\}@marley.uni-muenster.de%
}

\abstract{We report on a recently initiated project which aims at building a
  multi-layered parallel treebank of English and German. Particular attention
  is devoted to a dedicated predicate-argument layer which is used for
  aligning translationally equivalent sentences of the two languages. We
  describe both our conceptual decisions and aspects of their technical
  realisation. We discuss some selected problems and conclude with a few
  remarks on how this project relates to similar projects in the field.}

\begin{document}


\newcommand{\fuse}[0]{FuSe}
\newcommand{\dbliteral}[1]{\texttt{#1}}


\maketitleabstract

\section{Introduction}
\label{sec:intro}

Parallel corpora are widely accepted as a valuable data source for machine
translation and other research.  So far, however, the amount of linguistic
annotation in these corpora is limited, and particularly multilingual corpora
annotated with syntactic information are rare.  Our goal is to build a
treebank of aligned parallel\footnote{In accordance with the terminology
  suggested in \cite{Sinclair94}, we understand ``parallel'' to mean that
  the texts are translations of each other.} texts in English and German with
the following linguistic levels: \textsc{pos} tags, constituent structure,
functional relations and predicate-argument structure for each monolingual
subcorpus, plus an alignment layer to ``fuse'' the two -- hence our working
title for the treebank, \fuse{}, which additionally stands for
\emph{fu}nctional \emph{se}mantic annotation \cite{CyrusEtAl03p}.

We use the Europarl Corpus \cite{Koehn02}, which contains sentence-aligned
proceedings of the European parliament in eleven languages and thus offers
ample opportunity for extending the treebank at a later stage.\footnote{There
  are a few drawbacks to Europarl, such as its limited register and the fact
  that it is not easily discernible which language is the source language.
  However, we believe that at this stage the easy accessibility, the amount of
  preprocessing and particularly the lack of copyright restrictions make up
  for these disadvantages.} For syntactic and functional annotation we
basically adapt the \textsc{tiger} annotation scheme \cite{AlbertEtAl03},
making adjustments where we deem appropriate and changes which become
necessary when adapting to English an annotation scheme which was originally
developed for German.

The fusion of the language pair will take place on an alignment layer which
connects the predicate-argument layers of both monolingual subcorpora.  Only
the alignment layer is explicitly defined for a language pair rather than for
a single language. Apart from this layer, the subcorpora are monolingual
resources in their own right.

Although, eventually, the treebank will prove useful for several
fields of application, the most obvious one being machine translation, our
main motivation is to contribute to linguistic research.  The treebank will
serve as a resource for both monolingual and contrastive analyses.

\section{Reasons for Predicate-Argument Structure}
\label{sec:reasons}

In a parallel treebank, it is necessary to capture the translational
equivalence between two sentences. Our basic assumption is that this
equivalence can best be represented by means of a predicate-argument
structure. It is sometimes assumed that predicate-argument structure can be
derived or recovered from constituent structure or functional tags such as
subject and object.\footnote{See e.\,g.\ \cite{MarcusEtAl94}.} While it is
true that these annotations provide important heuristic clues for the
identification of predicates and arguments, predicate-argument structure goes
beyond the assignment of phrasal categories and grammatical functions, because
the grammatical category of predicates and consequently the grammatical
functions of their arguments can vary.

For instance, it is very common for an English verbal predicate to be
expressed by a nominalisation in German, as is the case in the \textsc{np}s in
\ref{ex:nominate} and \ref{ex:nominierung}, where the English verb
\emph{nominate} is translated as the German noun \emph{Nominierung}.

\ex. their automatic right to nominate a member of the European
Commission\footnote{Europarl:de-en/ep-00-02-15.al, 326. Note that throughout
  this paper, sentences are sometimes cited with irrelevant parts omitted.}
\label{ex:nominate}

\exg. ihr automatisches Recht auf Nominierung eines Mitglieds der
Europ\"{a}ischen Kommission\\
their automatic right on nomination of\_a member of\_the European
Commission\\
\label{ex:nominierung}

\begin{figure*}[p]
  \begin{center}
    \epsfig{file=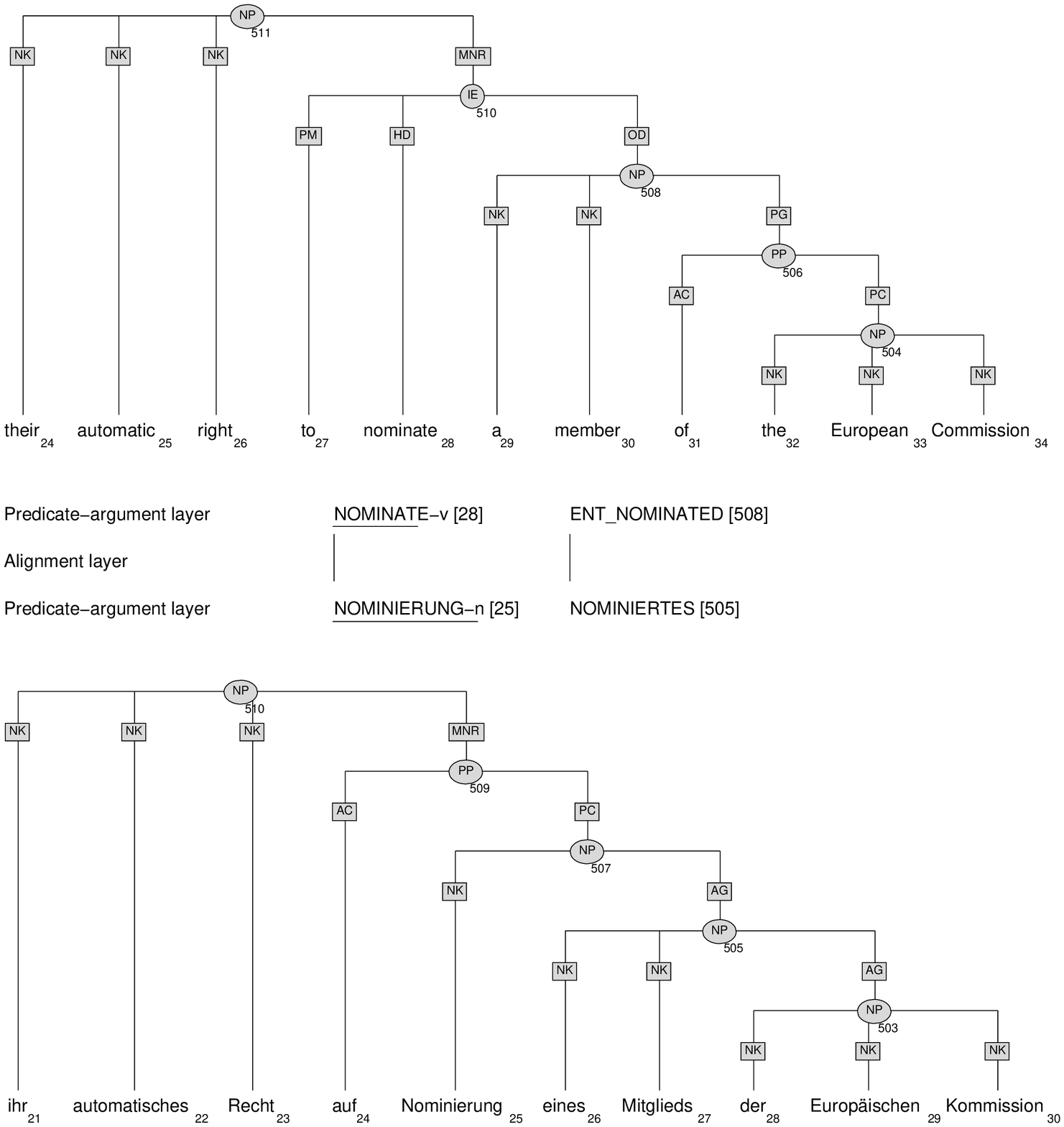,height=.55\textheight}
    \caption{Alignment of a verb/direct-object construction 
      with a noun/modifier construction}
    \label{fig:reason}
  \end{center}
\end{figure*}

The annotations of these noun phrases are shown in
Figure~\ref{fig:reason}.\footnote{All figures are at the end of the paper.} It
can be seen that the correspondence between \textsc{np}$_{508}$ and
\textsc{np}$_{505}$ cannot be inferred from the constituent structure, since
\textsc{np}$_{508}$ is an immediate constituent of an \textsc{ie} (``extended
infinitive'') while \textsc{np}$_{505}$ is deeply embedded in a \textsc{pp}.
Neither can the correspondence of \textsc{np}$_{508}$ and \textsc{np}$_{505}$
be inferred from their respective functional categories, since
\textsc{np}$_{508}$ is a direct object (\textsc{od}) while \textsc{np}$_{505}$
is a modifier (\textsc{ag}: ``genitive attribute'').  However, the resemblance
between these constituents becomes apparent when they are marked for their
argument status, because they both fulfill a similar role.

We have therefore chosen to represent predicate-argument structure on a
dedicated layer in our treebank in order to be able to capture the parallelism
between translations and to use it as the basis for alignment.

\section{Details of the Predicate-Argument Annotation}
\label{sec:details}

The predicate-argument structures used here consist solely of predicates and
their arguments. Although there is usually more than one predicate in a
sentence, no attempt is made to nest structures or to join the predications
logically in any way.\footnote{Since the predicate-argument structure is
  always bound to the constituent structure (see Section \ref{sec:bind}), it
  might well be possible to derive this information, e.\,g.\ through
  coordination structures and the hierarchical ordering of constituents.}  The
idea is to make the predicate-argument structure as rich as is necessary to be
able to align a sentence pair while keeping it as simple as possible so as not
to make it too difficult to annotate.  In the same vein, quantification,
negation, and other operators are not annotated. In short, the
predicate-argument structures are not supposed to capture the semantics of a
sentence exhaustively in an interlingua-like fashion.

\subsection{Predicates and Arguments}
\label{sec:pa}

In determining what a predicate is and how many there are in a sentence we
rely on a few assumptions that are of a heuristic nature. One of these
assumptions is that predicates are more likely to be expressed by tokens
belonging to some word classes than by tokens belonging to others.  Potential
predicate expressions in \fuse\ are verbs, deverbal adjectives and
nouns\footnote{For all non-verbal predicate expressions for which a
  derivationally related verbal expression exists it is assumed that they are
  deverbal derivations, etymological counter-evidence notwithstanding.} or
other adjectives and nouns which show a syntactic subcategorisation pattern.
The predicates are represented by the capitalised citation form of the lexical
item (e.\,g.\ \textsc{nominate}).  Homonymous or polysemous predicates are
differentiated by means of a disambiguator, predicates are assigned a class
based on their syntactic form, and derivationally related predicates form a
predicate group.

Arguments are given short intuitive role names (e.\,g.\ 
\textsc{ent\_nominated}) in order to facilitate the annotation process.  These
role names have to be used consistently only within a predicate group.  If,
for example, an argument of the predicate \textsc{nominate} has been assigned
the role \textsc{ent\_nominated} and the annotator encounters a comparable
role as argument to the predicate \textsc{nomination}, the same role name for
this argument has to be used.

Keeping the argument names consistent for all predicates within a group while
differentiating the predicates on the basis of syntactic form are
complementary principles, both of which are supposed to facilitate querying
the corpus.  The consistency of argument names within a group, for example,
enables the researcher to analyse paradigmatically all realisations of an
argument irrespective of the syntactic form of the predicate. At the same
time, the differentiation of predicates makes possible a syntagmatic analysis
of the differences of argument structures depending on the syntactic form of
the predicate.

\subsection{Binding Layer}
\label{sec:bind}

All elements of the predicate-argument structure must be bound to elements of
the phrasal structure (terminal or non-terminal nodes). These bindings are
stored in a dedicated binding layer between the constituent layer and the
predicate-argument layer.  

When an expected argument is absent on the phrasal level due to specific
syntactic constructions, the binding of the predicate is tagged accordingly,
thus accounting for the missing argument. For example, in passive
constructions like in Table \ref{tab:tagpred}, the predicate binding is tagged
as \dbliteral{pv}. Other common examples are imperative constructions.
Although information of this kind may possibly be derived from the constituent
structure, it is explicitly recorded in the binding layer as it has a direct
impact on the predicate-argument structure.

\begin{table}[htbp]
  \centering
  \begin{tabular*}{\columnwidth}{@{\extracolsep{\fill}}l@{\quad}cccc}
    \hline
    \emph{Sentence} & wenn & korrekt   & gedolmetscht & wurde          \\
    \emph{Gloss}    & if   & correctly & interpreted  & was            \\
                    &      &           & $\uparrow$   &                \\
    \emph{Binding}  &      &           & \dbliteral{pv}  &             \\
                    &      &           & $\vert$      &                \\
    \emph{Pred/Arg} &      &           & \textsc{dolmetschen} &        \\
    \hline
  \end{tabular*}
  \caption{Example of a tagged predicate binding
    (Europarl:de-en/ep-00-01-18.al, 2532)}
  \label{tab:tagpred}
\end{table}

Bindings of arguments may be tagged as well, an example for this being
object-control (cf. Table \ref{tab:tagarg}). To account for the deviant case
of the subject of the embedded clause in an object-control construction, the
binding of this argument is tagged (\dbliteral{oc-case}).  With this
information, a researcher or a machine learner will be able to ignore a
specific argument which might distort statistics on the phrasal realisations
of arguments.

The predicate binding is tagged as well to mark the entire object-control
construction (\dbliteral{oc}). This tagging enables the researcher to filter
out this specific predicate-argument structure, so as to ignore these
constructions completely.

\begin{table*}[htbp]
  \centering
  \begin{tabular*}{\textwidth}{@{\extracolsep{\fill}}l@{\qquad}ccccc}
    \hline
    \emph{Sentence} & 
    It was this which inspired & 
    us &
    to propose & 
    the same thing &
    with regard to state aid . 
    \\

    & 
    & 
    $\uparrow$ &
    $\uparrow$ &
    $\uparrow$ &
    \\

    \emph{Binding} &      
    &           
    \dbliteral{oc-case} & 
    \dbliteral{oc} &
    [] &
    \\

    & 
    & 
    $\vert$ &
    $\vert$ &
    $\vert$ &
    \\
    
    \emph{Pred/Arg} & 
    &
    \textsc{proposer} & 
    \textsc{propose} &
    \textsc{proposal} 
    \\

    \hline
  \end{tabular*}
  \caption{Example of tagged predicate and argument bindings
    (Europarl:de-en/ep-00-01-18.al, 237)}
  \label{tab:tagarg}
\end{table*}

Section \ref{sec:binding} will show that linking predicates or arguments to
constituents cannot always be achieved by binding them to a single node in the
constituent structure. In order to be flexible in this respect, the binding
layer allows for complex bindings, with more than one node of the constituent
structure to be included in and sub-nodes to be explicitly excluded from a
binding to a predicate or argument.\footnote{See the database documentation
  \cite{Feddes04} for a more detailed description of this mechanism.}

\subsection{Alignment Layer}
\label{sec:align}

On the alignment layer, the elements of a pair of predicate-argument
structures are aligned with each other.  Arguments are aligned on the basis of
corresponding roles within the predications.  Comparable to the tags used in
the binding layer that account for specific constructions (see Section
\ref{sec:bind}), the alignments may also be tagged with further information.
This becomes necessary when the predications are incompatible in some way.
Section \ref{sec:incompatible} will give examples.

If there is no corresponding predicate-argument structure in the other
language or if an argument within a structure does not have a counterpart in
the other language, there will simply be no alignment. Section
\ref{sec:modality} provides an example where a predication is left dangling.

Table \ref{tab:layers} gives an overview of the annotation layers as described
in this section.

\begin{table}[htbp]
  \centering
  \begin{tabular*}{\columnwidth}{l@{\extracolsep{\fill}}l}
    \hline
    \emph{Layer} & \emph{Function}\\
    \hline
    Phrasal & constituent structure of language A\\
    Binding & binding $\downarrow$ predicates/arguments to $\uparrow$ nodes\\
    \textsc{pa} & predicate-argument structures\\
    \hline
    Alignment & aligning $\updownarrow$ predicates and arguments \\
    \hline
    \textsc{pa}& predicate-argument structures\\
    Binding & binding $\uparrow$ predicates/arguments to $\downarrow$ nodes\\
    Phrasal & constituent structure of language B\\
    \hline
  \end{tabular*}
  \caption{The layers of the predicate-argument annotation}
  \label{tab:layers}
\end{table}

\section{Problematic Cases}
\label{sec:problems}

In this section we will elaborate on some problematic cases of
predicate-argument annotation which we have encountered so far, some of them
particular to the annotation and alignment of predicate-argument structures
for a language pair.

\subsection{Binding Predicate-Argument Structure to Constituent Structure}
\label{sec:binding}

\begin{figure*}[p]
  \begin{center}
    \epsfig{file=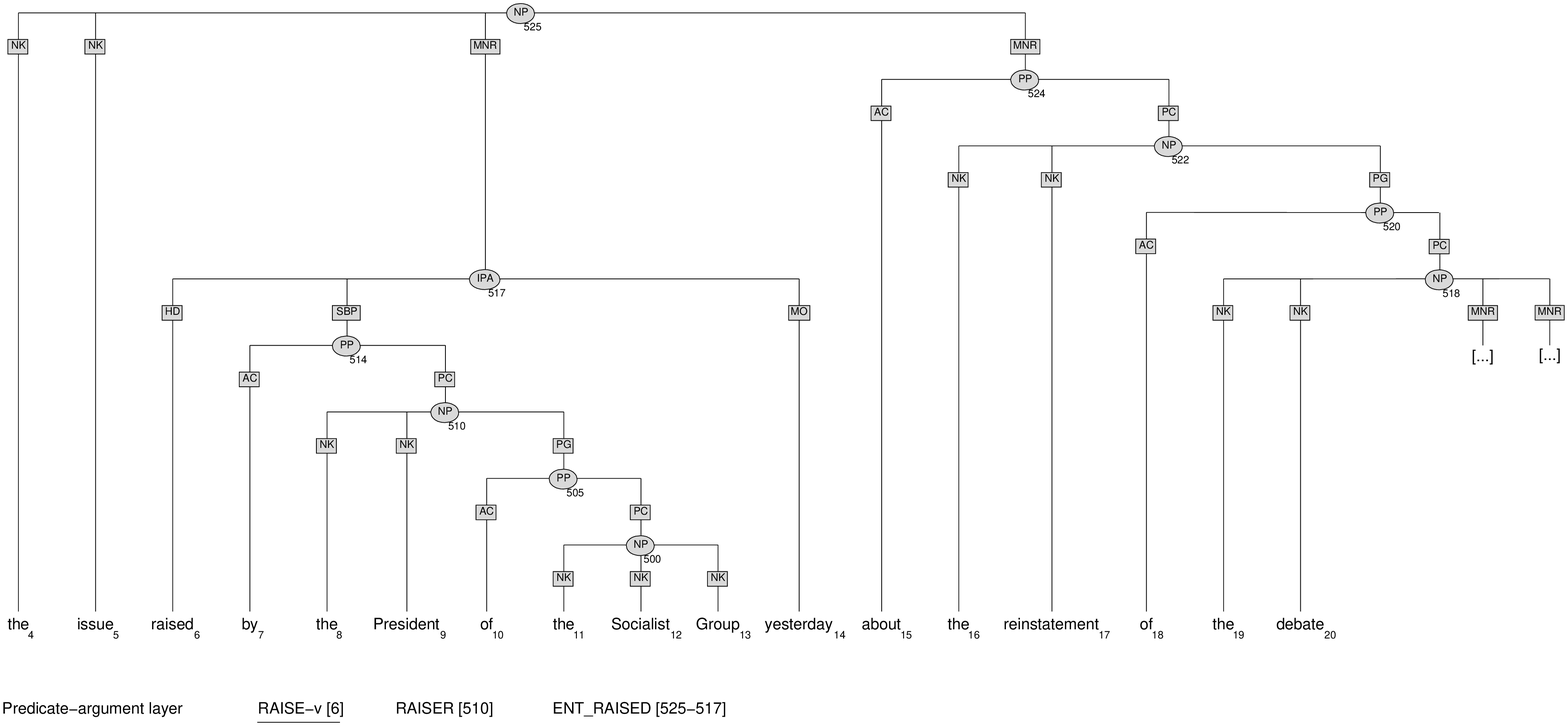,width=\textwidth}
    \caption{Complex constituent binding of an argument}
    \label{fig:binding}
  \end{center}
\end{figure*}

It was mentioned in Section \ref{sec:details} that all predicates and
arguments must be bound to either terminal or non-terminal nodes in the
constituent structure. However, this is not always possible since in some
cases there is no direct correspondence between argument roles and
constituents. For instance, this problem occurs whenever a noun is
postmodified by a participle clause: in Figure~\ref{fig:binding}, the argument
role \textsc{ent\_raised} of the predicate \textsc{raise} is realised by
\textsc{np}$_{525}$, but the participle clause (\textsc{ipa}$_{517}$)
containing the predicate (\emph{raised}$_6$) needs to be excluded, because not
excluding it would lead to recursion.  Consequently, there is no simple way to
link the argument role to its realisation in the tree.

In these cases we link the argument role to the appropriate phrase (here:
\textsc{np}$_{525}$) and prune out the constituent that contains the predicate
(\textsc{ipa}$_{517}$; see Section \ref{sec:bind} for this mechanism), which
results in a discontinuous argument realisation.

\subsection{Coping with Modality}
\label{sec:modality}

Generally, modal verbs are not considered to be predicates and are
consequently not included in our predicate-argument database. This can cause a
problem when a verbal predicate that is modified by a modal auxiliary in L1
\ref{ex:harmonise} is represented by a deverbal noun in the corresponding
sentence in L2 \ref{ex:harmonisierung}.

\ex. The laws against racism must be
harmonised.\footnote{Europarl:de-en/ep-00-01-19.al, 489.}
\label{ex:harmonise}

\exg.  Die Harmonisierung der Rechtsvorschriften gegen den Rassismus ist
dringend erforderlich.\\
The harmonisation of\_the laws against the racism is urgently necessary.\\
\label{ex:harmonisierung}

\begin{figure*}[p]
  \begin{center}
    \epsfig{file=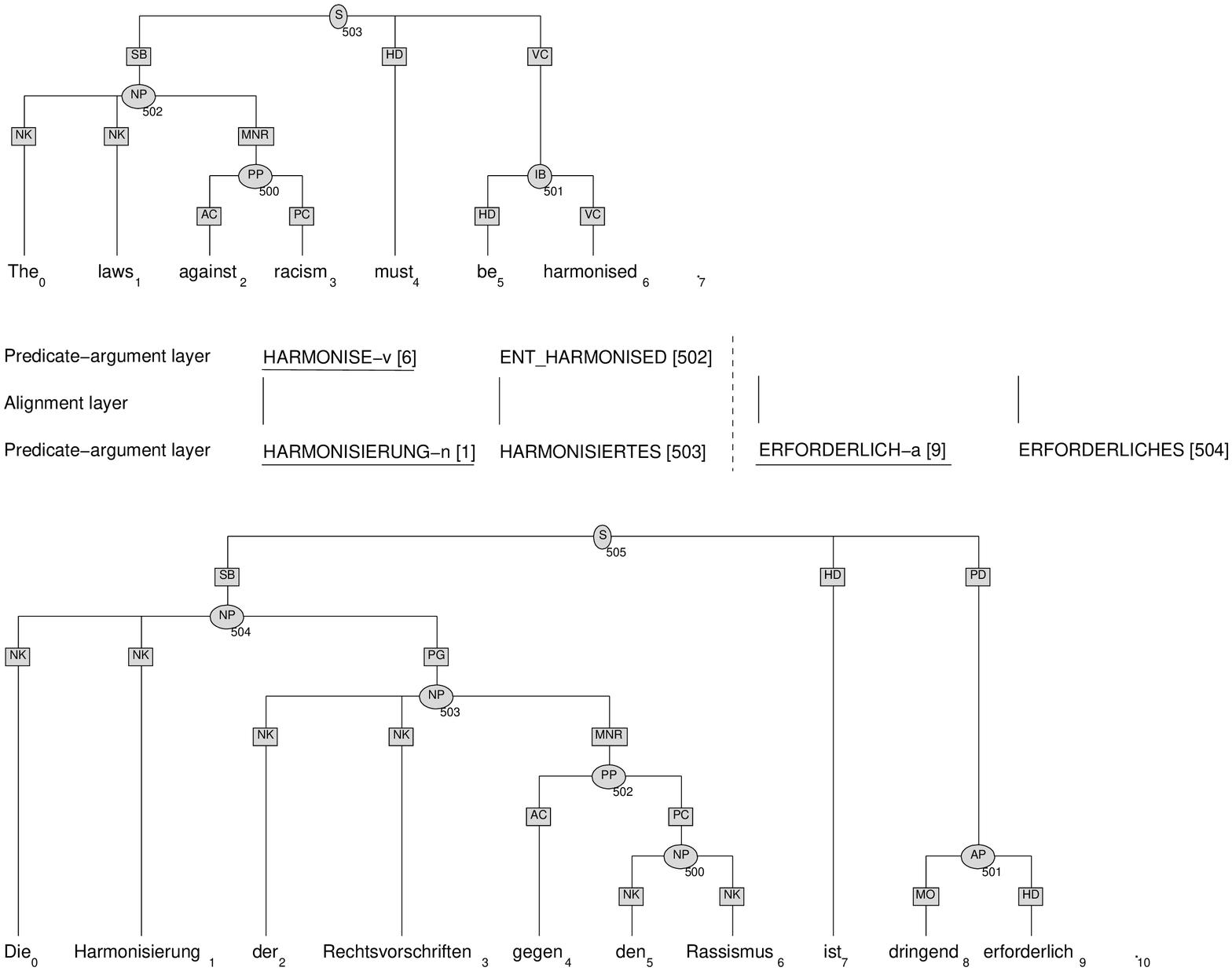,height=.4\textheight}
    \caption{Modality}
    \label{fig:modality}
  \end{center}
\end{figure*}

This can be illustrated by Figure~\ref{fig:modality}: the realisation of the
verbal predicate \textsc{harmonise} (\emph{harmonised}$_6$) is modified by the
modal auxiliary \emph{must}$_4$. In the German sentence, the nominal predicate
\textsc{harmonisierung} (\emph{Harmonisierung}$_1$) is used. Here, the
modality is expressed by a predicate of its own, namely \textsc{erforderlich}
(\emph{erforderlich}$_9$, `necessary'). This second predicate does not
correspond to any predicate in the English sentence.

It would be an easy way out to resort to annotating modal auxiliaries as if
they were full verbs and consequently predicates, but we have opted against
this makeshift solution. One has to keep in mind that the predicate-argument
annotation is done monolingually and only later serves as the basis for
alignment. It should not be assumed that the corresponding equivalent is known
to the annotator during the annotation process. Even though the way a sentence
is expressed in another language can give valuable insights into its structure
and meaning, this should not go so far as to change the way the original
language is annotated.  This is particularly true since the idea behind the
FuSe treebank is that it is in principle extendable and may well include
languages other than English and German in the future. As it cannot be
foretold what phenomena will be encountered once further languages are
added, the decisions as to what is annotated and what is not should not be
guided by cross linguistic considerations.

Thus, the simple fact alone that a predication in one language does not
correspond to a predication in another should not induce one to alter the
annotation praxis so as to make the two versions more compatible with each
other. Modality, in particular, can be expressed in a variety of ways, and
just because one of them is the realisation as a predicative adjective does
not make, say, a modal adverbial like \emph{certainly} a predicate. The same
argumentation holds for modal auxiliaries.

\subsection{Incompatible Predications}
\label{sec:incompatible}

\begin{figure*}[p]
  \begin{center}
    \epsfig{file=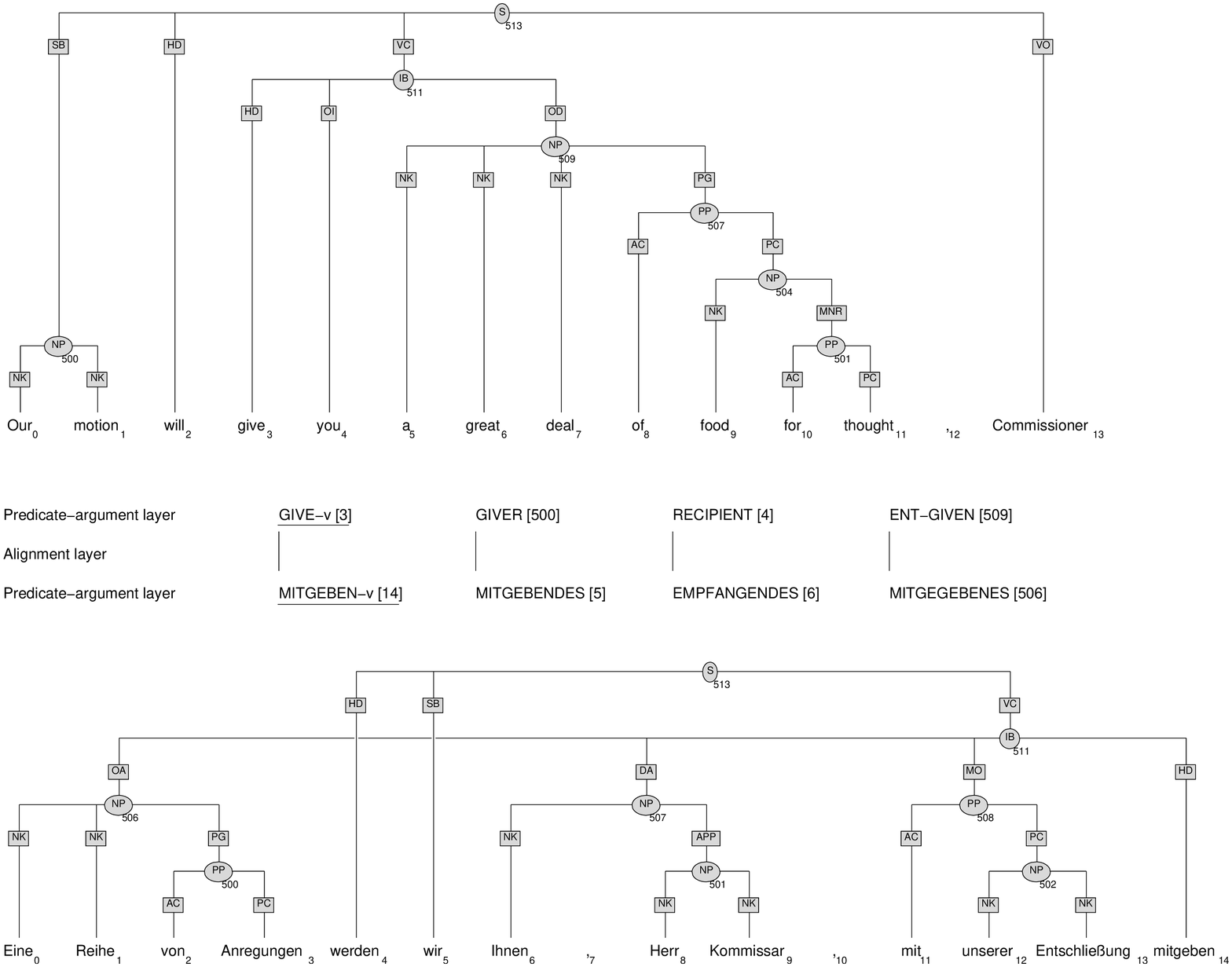,height=.45\textheight}
    \caption{Incompatible predications}
    \label{fig:incompatible}
  \end{center}
\end{figure*}

Sometimes, the predications in two corresponding sentences express
approximately the same idea but are otherwise incompatible with each other.
This can be demonstrated with sentences \ref{ex:motion} and \ref{ex:mitgeben},
the annotation, argument structure and alignment of which are illustrated in
Figure~\ref{fig:incompatible}.

\ex. Our motion will give you a great deal of food for thought,
Commissioner\footnote{Europarl:de-en/ep-00-01-18.al, 53.}
\label{ex:motion}

\exg. Eine Reihe von Anregungen werden wir Ihnen, Herr Kommissar, mit unserer
Entschlie{\ss}ung mitgeben\\
A row of suggestions will we you, Mr. Commissioner, with our resolution give\\
\label{ex:mitgeben}

The incompatibility results from the fact that, while the predicates
\textsc{give} and \textsc{mitgeben} are roughly equivalent in meaning, the two
sentences are organised differently with regard to their information
structure. This has caused the two corresponding argument roles of
\textsc{giver} and \textsc{mitgeber} to be realised by two incompatible
expressions representing different referents (\textsc{np}$_{500}$ vs.
\emph{wir}$_5$).  The English version is somewhat metaphorical in that, unlike
in the German sentence, there is no animate entity in this agent-like argument
position. The actual agent is not realised as such and can only be identified
by a process of inference based on the presence of the possessive pronoun
\emph{our}$_0$. To complicate matters even further, the translational
equivalent of \textsc{np}$_{500}$ (i.\,e.\ the constituent realising the
English \textsc{giver}), is not even an argument in the German sentence
(\textsc{pp}$_{508}$).

Consequently, it seems impossible to reach a satisfactory alignment in this
case: either two arguments with the same role but different meanings would
have to be aligned, or else the alignment would rely solely on translational
equivalence, which would reduce to absurdity our reasons for including
predicate-argument structure.

We solve the problem as follows: since cases like this are at the same time
potentially interesting for contrastive analyses and a hazard for applications
using the treebank for automatic learning, we keep up the alignment on the
basis of argument roles but tag the alignment (see Section \ref{sec:align})
between the arguments in question and thus mark them as being incompatible
(\dbliteral{incomp}) with each other. This enables the interested researcher
to formulate explicit searches for this alignment type while making it
possible for applications to skip these cases if this is preferred.

Sentences \ref{ex:inapplicable} and \ref{ex:anwendbar} are a second case where
we make use of the possiblilty to tag the alignment. Here, the adjectival
predicate \textsc{inapplicable} in \ref{ex:inapplicable} is represented by the
negated predicate \textsc{anwendbar} (`applicable') in the German counterpart
\ref{ex:anwendbar}. 

\ex. the Directive is inapplicable in Denmark
\footnote{Europarl:de-en/ep-00-01-18.al, 2522.}
\label{ex:inapplicable}

\exg. die Richtlinie ist in D\"{a}nemark nicht anwendbar\\
the Directive is in Denmark not applicable\\
\label{ex:anwendbar}

Since whether or not a predicate is negated does not alter its argument
structure we do not annotate negation (see Section \ref{sec:details}). As this
leads to an alignment of predicates with opposite meanings, we tag the
alignment between the two predicates as \dbliteral{abs-opp} (``absolute
opposites''). In theory, this method could also be applied to cases where a
predicate is translated by its relational opposite (e.\,g.\ \emph{buy} vs.
\emph{sell}). So far, however, we have not yet come across this type of
translation in our data. It will be interesting to discover what types of
incompatibility will come to light as the annotation proceeds.

\section{Database Structure and Tools}
\label{sec:db}

We use \textsc{Annotate} \cite{Plaehn98b} for the semi-automatic assignment
\cite{Brants99} of \textsc{pos} tags, hierarchical structure, phrasal and
functional tags. \textsc{Annotate} stores all annotations in a relational
database.\footnote{For details about the \textsc{Annotate} database structure
  see \cite{Plaehn98}.} To stay consistent with this approach we have
developed an extension to the \textsc{Annotate} database structure to model
the predicate-argument layer and the binding layer.

Due to the monolingual nature of the \textsc{Annotate} database structure, the
alignment layer (Section \ref{sec:align}) cannot be incorporated into it.
Hence, additonal types of databases are needed. For each language pair
(currently, English and German), an alignment database is defined which
represents the alignment layer, thus fusing two extended \textsc{Annotate}
databases.  Additionally, an administrative database is needed to define sets
of two \textsc{Annotate} databases and one alignment database. The final
parallel treebank will be represented by the union of these sets
\cite{Feddes04}.

While annotators use \textsc{Annotate} to enter phrasal and functional
structure comfortably, the predicate-argument structures and alignments are
currently entered into a structured text file which is then imported into the
database. A graphical annotation tool for these layers is under development.
It will make binding the predicate-argument structure to the constituent
structure easier for the annotators and suggest argument roles based on
previous decisions.

\section{Relation to Other Projects and Outlook}
\label{sec:relation}

This section will show briefly how our approach relates to other projects
annotating some kind of predicate-argument structure, such as PropBank
\cite{PalmerEtAl03} and FrameNet \cite{JohnsonEtAl03}, and how the alignment
structures of the parallel treebank make up for certain drawbacks of our
annotation scheme.

Since our annotation of predicates and their arguments is not a means in
itself but to the end of aligning constituents of a parallel treebank, it is
kept deliberately simple.  It resembles the mnemonic descriptors clarifying
the numbered arguments in the PropBank framesets.  We do not, however, attempt
any generalisation whatsoever: neither do we organise our predicates in
frames, as is done by FrameNet and adopted by \textsc{salsa}
\cite{ErkEtAl03a}, nor do we follow the Levin classes \cite{Levin93}, as is
done in the PropBank project.

Some problems we encounter with our simple scheme could be avoided with a
deeper predicate-argument structure. As the first example in Section
\ref{sec:incompatible} shows, predications which are incompatible in our
scheme need not be incompatible in a FrameNet-like scheme: if the argument
roles were deeper than our intuitive role names, i.\,e., if \emph{our motion}
in example \ref{ex:motion} were not a \textsc{giver} but, e.\,g., a
\textsc{cause}, the incompatibility with the corresponding structure in
\ref{ex:mitgeben} would not arise.

There are several reasons for us to stick to our simple approach.  For one
thing, a more complex scheme would make the annotation more susceptible to
inconsistencies. Secondly, transferring the approaches mentioned above to
other languages than English is not a straightforward matter.  While this
seems to be working quite well for the FrameNet frames \cite{ErkEtAl03a},
Levin's verb classes are inherently English and cannot be directly applied to
German.  In a later stage of the project, it might be possible to work through
the predicate-argument database and map our very specific scheme to a more
general one, e.\,g. by assigning each predicate to a frame and each argument
to a frame element. However, other studies show that mapping one scheme onto
another is far from trivial \cite{HajicovaKucerova02}, and quite a lot of
manual work will presumably be necessary.

\begin{figure*}[p]
  \begin{center}
    \epsfig{file=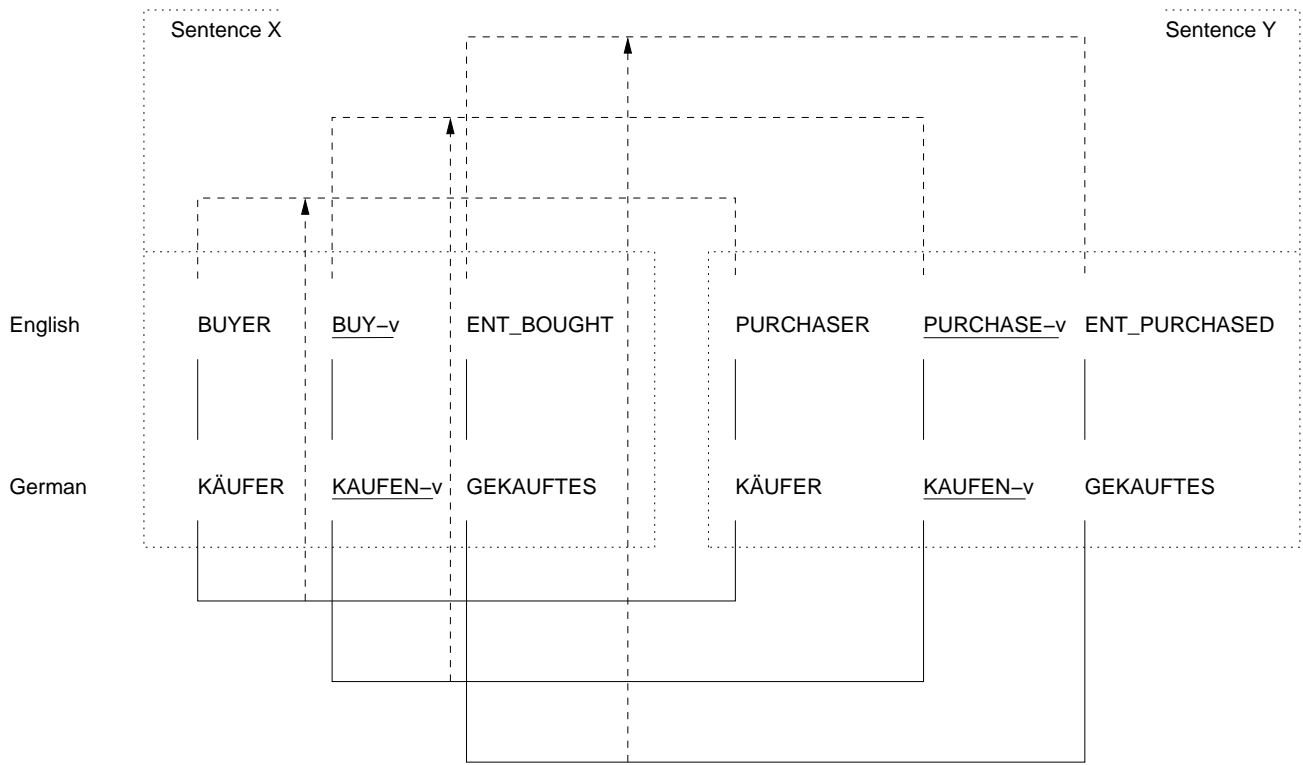,width=\textwidth}
    \caption{Deriving predicate clusters by exploiting alignment
      structures}
    \label{fig:mapping}
  \end{center}
\end{figure*}

Finally, we believe it is possible to exploit the corpus as a parallel lexical
resource to see how different predicates can be clustered automatically by
analysing their mappings in the other language. Figure \ref{fig:mapping}
sketches the general idea.  Suppose that in the English sub-corpus, two
predicate-argument structures have different predicates (\textsc{buy} and
\textsc{purchase}) which subcategorise for comparable arguments and express
the same concept. In a FrameNet-like annotation, these predicates would be
instantiations of the same frame (e.\,g.\ \textsc{commercial\_transaction}).
In our scheme, neither are these predicates grouped in any way, nor do the
comparable arguments get the same role names.

However, it is well conceivable that both predicates are translated
identically in the corresponding German structures (e.\,g.\ by \textsc{kaufen}
`buy'). Since predicates and arguments are aligned to each other, the
comparability of the predicates (\textsc{buy} -- \textsc{purchase}) and their
arguments (\textsc{buyer} -- \textsc{purchaser} and \textsc{ent\_bought} --
\textsc{ent\_purchased}) can be derived (cf.\ the dashed lines). It will then
be instructive to investigate how these clusters compare to FrameNet frames
and to explore to what extent such a data-driven approach to frame semantics
is feasible.


\bibliographystyle{lrec2000}
\bibliography{paper_bib}


\end{document}